%% file: main.tex
\documentclass[letterpaper,journal]{IEEEtran}

\input{math_commands.tex}

\usepackage{graphicx}
\usepackage[caption=false,font=footnotesize]{subfig}
\usepackage[table]{xcolor}
\usepackage{multirow}
\usepackage{enumitem}
\usepackage{placeins}
\usepackage{booktabs}
\usepackage{tabularx}
\usepackage{stfloats}
\usepackage[nocompress]{cite}
\usepackage{url}
\usepackage{pifont}
\usepackage{tikz}
\usepackage[hidelinks]{hyperref}

\title{
HiMEM: Hippocampus-Inspired Memory Consolidation, Pruning, and Composition for Vision-Language-Action Models in 3D Scenes
}

\author{Boran~Zhao, Yanbin~Hu, Jun~Ye, Zhenxian~Hu, Yuxin~Zhai,
Jin~Cui, and Pengju~Ren\IEEEauthorrefmark{1}~%
\thanks{Boran Zhao, Yanbin Hu, Jun Ye, Zhenxian Hu, and Yuxin Zhai
are with the School of Software Engineering, Xi'an Jiaotong University,
Xi'an, China.}%
\thanks{Jin Cui and Pengju Ren are with the School of Artificial
Intelligence, Xi'an Jiaotong University, Xi'an, China.}%
\thanks{All authors are also with the State Key Laboratory of Human-Machine Hybrid Augmented Intelligence, Institute of Artificial Intelligence and Robotics, Xi'an Jiaotong University,
Xi'an, China.}%
\thanks{\IEEEauthorrefmark{1}Corresponding author: Pengju Ren
(e-mail: pengjuren@xjtu.edu.cn).}}

\begin{document}
\maketitle

\begin{abstract}


Memory plays a crucial role in long-horizon task execution and cross-task generalization for Vision-Language-Action (VLA) models operating in complex 3D scenes. However, existing VLA memory mechanisms suffer from memory noise, particularly in complex 3D scenes, as stored episodic memories contain extensive redundant details about scene layouts, object configurations, and robot-specific states, which obscure generalizable manipulation principles and hinder cross-task transfer. To address this challenge, we first introduce a hippocampus-inspired memory consolidation mechanism that synthesizes extensive episodic experiences into a compact set of essential semantic memories, enabling VLA models to derive generalizable principles from practical experience. Second, we propose an episodic memory pruning mechanism that removes redundant episodic memories whose contents are captured by consolidated semantic memories, thereby increasing effective memory capacity and improving long-horizon task execution. Finally, to handle complex task scenarios, we develop a flexible memory composition mechanism that jointly leverages multiple episodic and semantic memories to enhance cross-scenario generalization. Experiments on LIBERO, LIBERO-Plus demonstrate the effectiveness of HiMEM across different VLA backbones, improving the average success rate of $\pi_0$ from 91.88\% to 98.20\% on LIBERO.



\end{abstract}


\bstctlcite{IEEE:BSTcontrol}

\input{sections/01_intro}
\input{sections/02_related}
\input{sections/03_method}
\input{sections/04_experiments}

\input{sections/05_limitation_and_conclusion}

\FloatBarrier


\bibliographystyle{IEEEtran}
\bibliography{refs}

\begin{IEEEbiography}[{\includegraphics[width=1in,height=1.25in,clip,keepaspectratio]{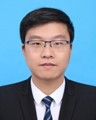}}]{Boran Zhao} is an assistant professor with the School of Software Engineering, Xi’an Jiaotong University. He received the Bachelor degree from Jiangnan University in 2015, the Master degree from Xidian University in 2018, and the Ph.D degree from Xi’an Jiaotong University in 2023. His current research interests are Brain inspired AI and Embodied Intelligence.
\end{IEEEbiography}

\begin{IEEEbiography}[{\includegraphics[width=1in,height=1.25in,clip,keepaspectratio]{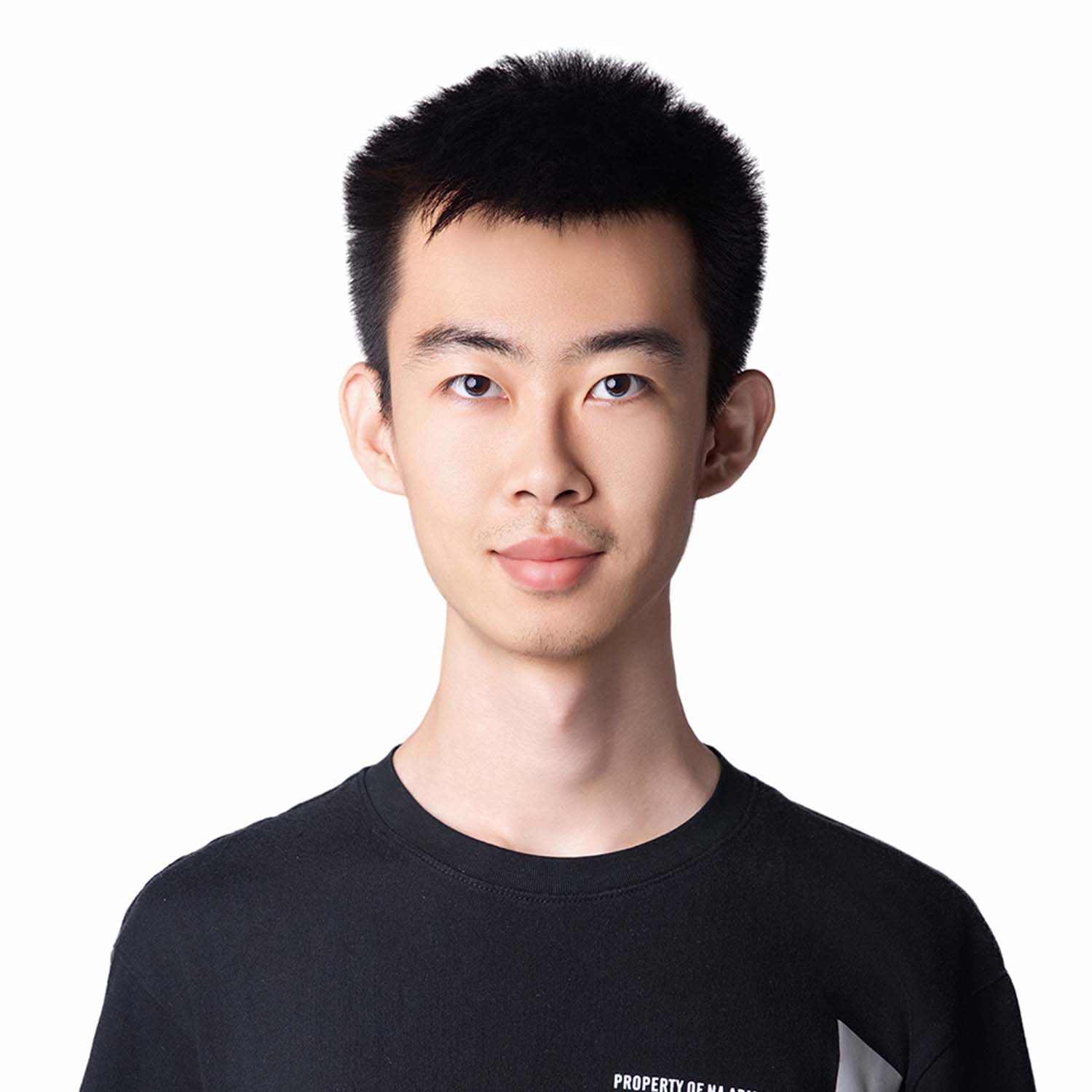}}]{Yanbin~Hu}
is a student in the School of Software Engineering, Xi’an Jiaotong University. His research interest is Embodied Intelligence.
\end{IEEEbiography}

\begin{IEEEbiography}[{\includegraphics[width=1in,height=1.25in,clip,keepaspectratio]{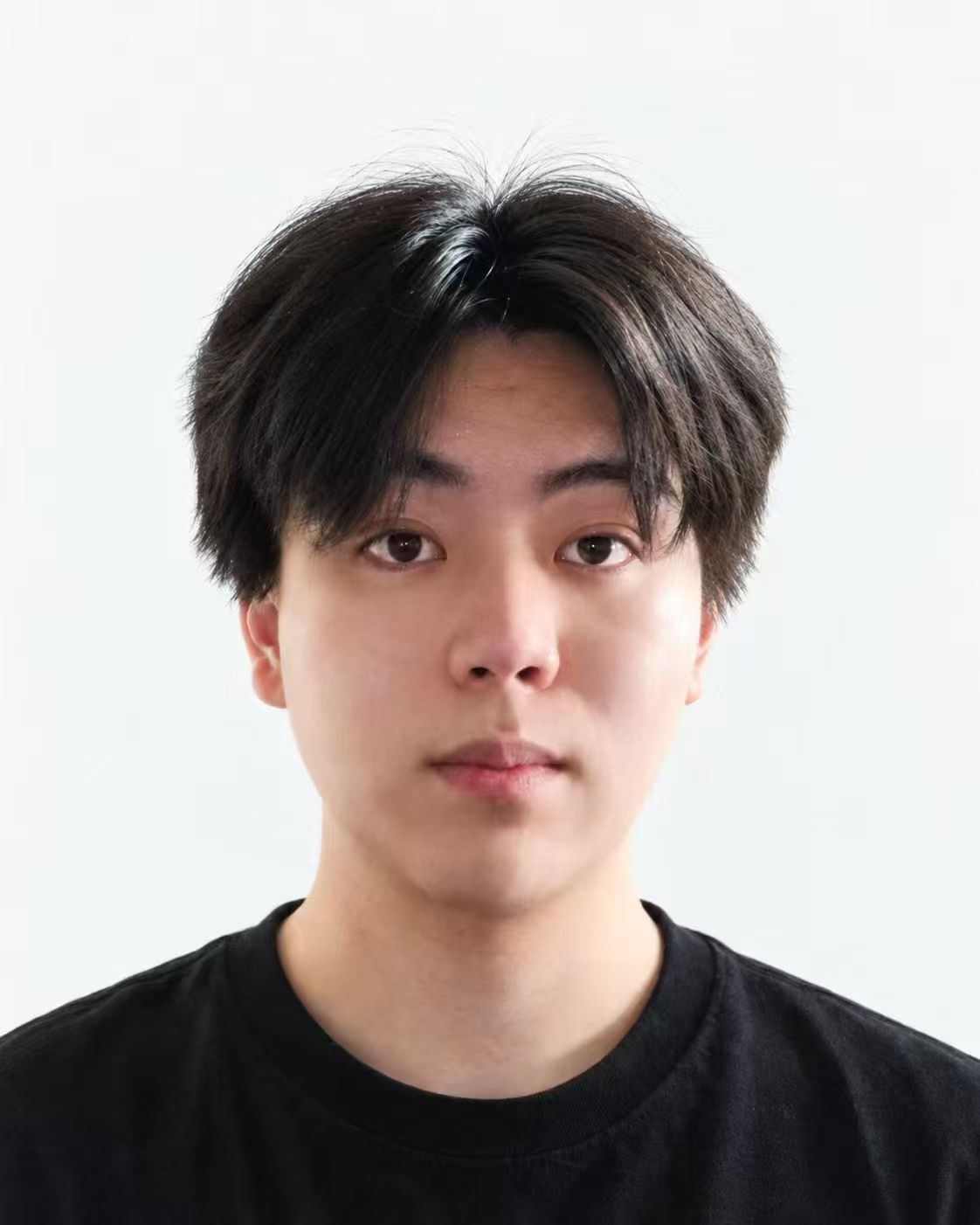}}]{Jun~Ye}
is a student in the School of Software Engineering, Xi’an Jiaotong University. His research interest is Embodied Intelligence.
\end{IEEEbiography}

\begin{IEEEbiography}[{\includegraphics[width=1in,height=1.25in,clip,keepaspectratio]{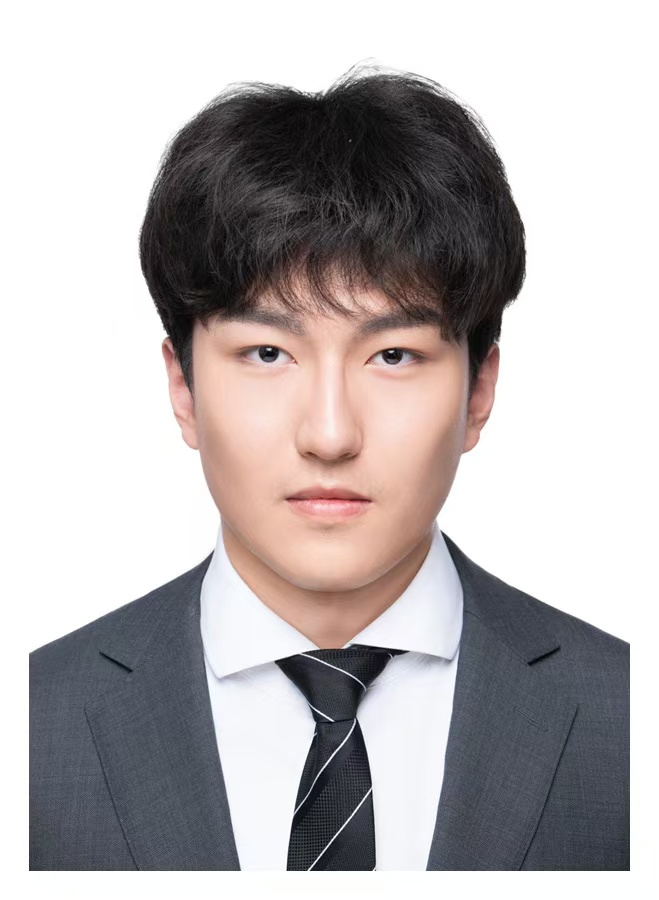}}]{Zhenxian~Hu}
is a student in the School of Software Engineering, Xi’an Jiaotong University. His research interest is Embodied Intelligence.
\end{IEEEbiography}

\begin{IEEEbiography}[{\includegraphics[width=1in,height=1.25in,clip,keepaspectratio]{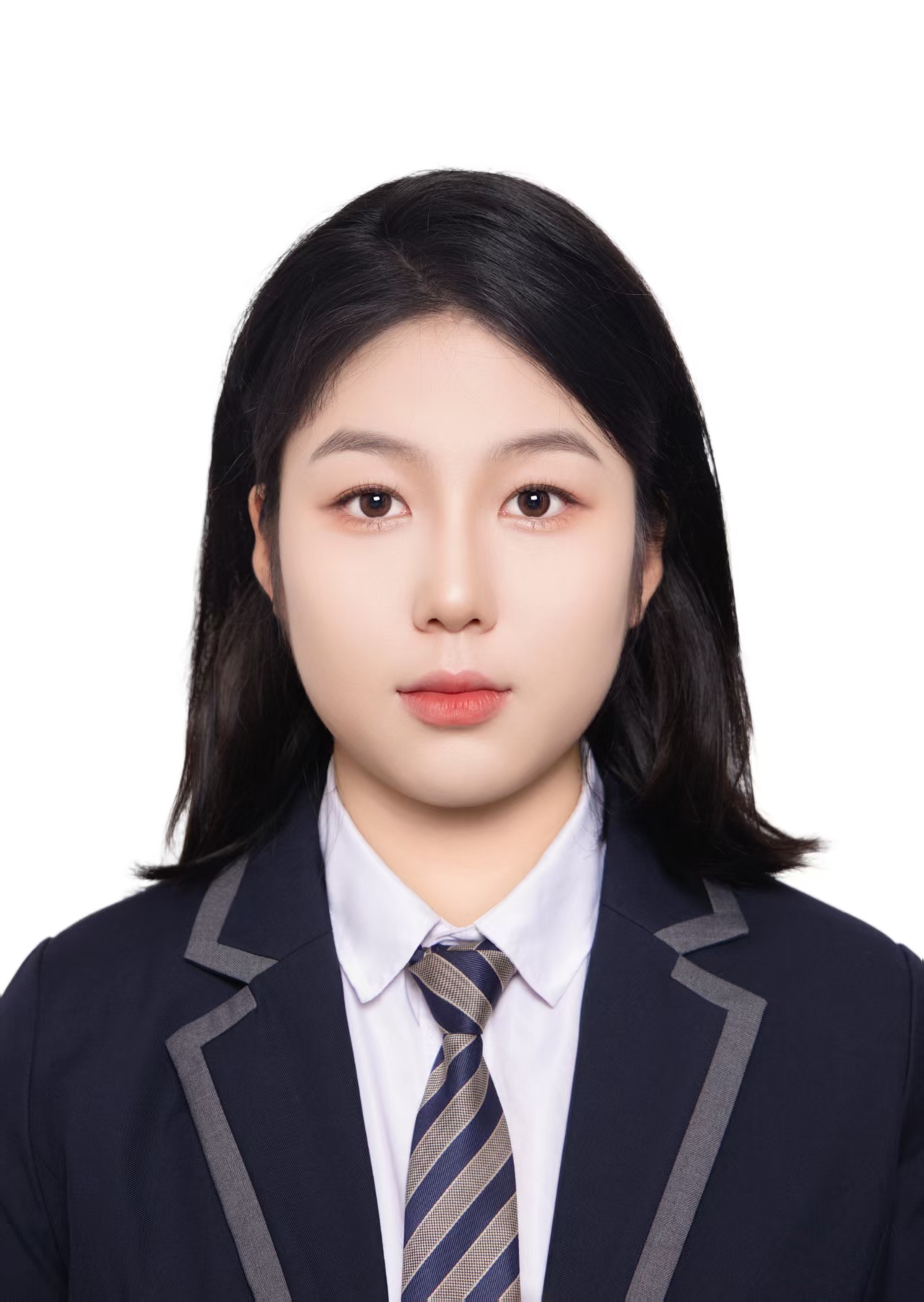}}]{Yuxin~Zhai}
is a student in the School of Software Engineering, Xi’an Jiaotong University. His research interest is Embodied Intelligence.
\end{IEEEbiography}

\begin{IEEEbiography}[{\includegraphics[width=1in,height=1.25in,clip,keepaspectratio]{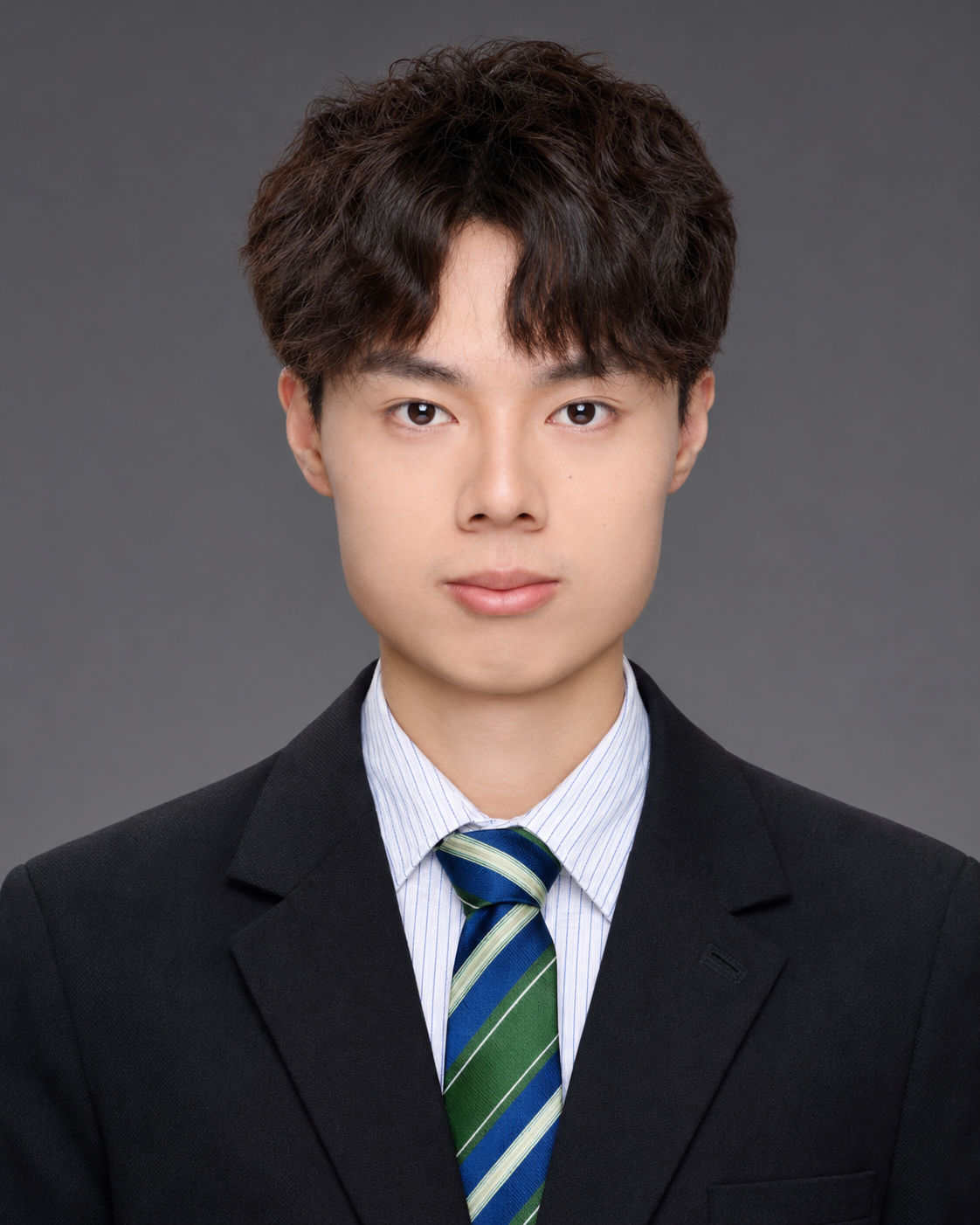}}]{Jin~Cui}
is a student in the School of Artificial Intelligence, Xi’an Jiaotong University. His research interest is Embodied Intelligence.
\end{IEEEbiography}

\begin{IEEEbiography}[{\includegraphics[width=1in,height=1.25in,clip,keepaspectratio]{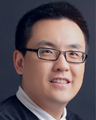}}]{Pengju Ren}
is a professor in the College of Artificial Intelligence, Xi’an Jiaotong University. He received Bachelor and PhD degree from Xi’an Jiaotong University in 2004 and 2012, respectively. During 2009 to 2011, he was a visiting PhD student in Computer Science and Artificial Intelligence Laboratory, Massachusetts Institute of Technology (MIT). His current research interests include Brain inspired AI and Embodied Intelligence. He is a member of the IEEE.
\end{IEEEbiography}

\end{document}

%% file: math_commands.tex
\usepackage{amsmath,amsfonts,bm}

\def\eqref#1{equation~\ref{#1}}

\def\1{\bm{1}}

\DeclareMathAlphabet{\mathsfit}{\encodingdefault}{\sfdefault}{m}{sl}
\SetMathAlphabet{\mathsfit}{bold}{\encodingdefault}{\sfdefault}{bx}{n}



%% file: sections/01_intro.tex
\section{Introduction}
\label{sec:introduction}
Vision-Language-Action (VLA) models have made substantial progress in robotic manipulation of 3D environments by unifying visual perception, language understanding, and action generation\cite{rt2,openvla,pi0}. However, long-horizon robotic manipulation typically involve multiple interdependent manipulation steps, requiring robots to continuously track environmental states and retain historical interaction information \cite{wang2026sliding,dinh2026vision}. Therefore, effective long-term memory mechanisms are crucial for improving the long-horizon robotic manipulation and cross-task generalization capabilities of VLA models.

\begin{figure}[ht!]
\centering
\includegraphics[width=\linewidth]{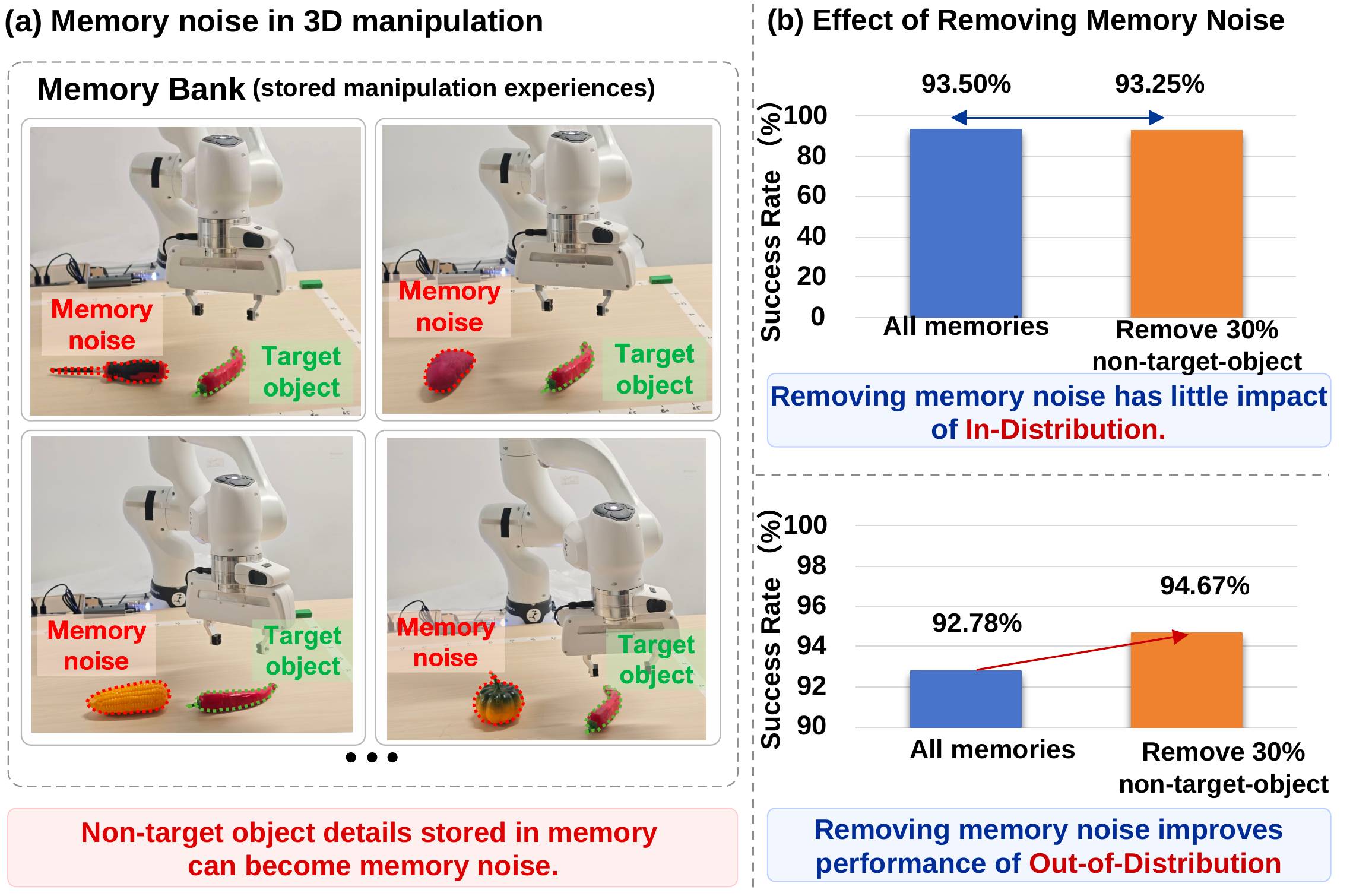}
\caption{\textbf{Illustration of memory noise and its effects on 3D manipulation.} (a) Stored manipulation experiences for the same target object and subgoal contain varying non-target object details, introducing memory redundancy. (b) Using ECHO~\cite{echo} for manipulation execution, we group memories by target and subgoal and randomly remove 30\% of redundant entries differing only in non-target object details while preserving task-relevant experiences. On LIBERO-Long, in-distribution success rates decrease slightly from 93.50\% to 93.25\%, whereas out-of-distribution success rates improve from 92.78\% to 94.67\%. These results suggest that removing redundant memory details can improve out-of-distribution generalization while largely preserving in-distribution performance.}

\label{fig:mem_noise}
\end{figure}

Most existing long-term memory mechanisms resemble episodic memory in the human brain, retaining extensive scene layouts, object states, and action outcomes from specific interactions, thereby providing rich information for robots to reuse historical experiences \cite{memoryvla,mem,map-vla,dejavu,memer}. However, especially in complex 3D scenes, these episodic memory records also contain substantial details that are irrelevant to the current task or specific to particular scenarios (i.e., memory noise, as shown in Fig.~\ref{fig:mem_noise}) and lack explicit guidance of generalizable manipulation patterns, resulting in limited transferability of experience across tasks \cite{echo}.


In contrast, humans possess an efficient memory system for interacting with 3D environments, which provides inspiration for addressing the aforementioned problem.
Specifically, complementary learning systems (CLS) theory \cite{mcclelland1995there} suggests that the hippocampus rapidly encodes and retains specific experiences, whereas the neocortex gradually develops generalizable semantic knowledge by integrating information across experiences. This division of labor enables the memory system to preserve information about specific events while identifying shared patterns across different experiences. 

Inspired by above mechanism, we propose HiMEM, a long-term memory framework for VLA models in 3D scenes. \ding{182}~First, HiMEM introduces a meta-semantic memory consolidation method inspired by the complementary mechanisms of the hippocampus and neocortex. The hippocampus-inspired module identifies recurring manipulation patterns and their applicable conditions across different experiences. The neocortex-inspired module then consolidates these patterns into compact meta-semantic memories that represent environmental and object conditions, reusable manipulation strategies, and their success probabilities, enabling VLA models to derive a compact set of abstract guiding principles from extensive concrete manipulation experiences. \ding{183}~Second, we propose an episodic memory pruning mechanism that partially removes the corresponding episodic memories once relevant semantic memories have been established, thereby increasing the memory capacity limit. \ding{184}~Finally, to address complex task scenarios, we introduce a mechanism for flexibly combining and utilizing multiple episodic and semantic memories. Specifically, meta-semantic memories provide manipulation patterns with conditional applicability, while episodic memories supplement information about specific manipulation procedures and state transitions. By incorporating visual observations of the current task, task instructions, and characteristics of the manipulated objects, the proposed mechanism flexibly combines memories at different levels of abstraction (i.e., episodic and semantic memories) to enable cross-scenario generalization in long-horizon tasks.
Experimental results demonstrate that, across the four standard LIBERO suites, our method improves the average success rates of three mainstream memory-free VLA models, $\pi_0$, $\pi_{0.5}$, and OpenVLA-OFT, by 6.32, 3.65, and 2.15 percentage points, respectively. Compared with ECHO~\cite{echo}, HiMEM further improves the average success rates by 0.90, 1.17, and 0.57 percentage points on $\pi_0$, $\pi_{0.5}$, and OpenVLA-OFT, respectively.

\begin{table*}[t]
\centering
\caption{Method comparison of this work and our previous conference version.}
\label{tab:method_comparison}
\renewcommand{\arraystretch}{1.35}
\setlength{\tabcolsep}{6pt}

\begin{tabularx}{\textwidth}{
    |>{\centering\arraybackslash}p{1.8cm}
    |>{\centering\arraybackslash}X
    |>{\centering\arraybackslash}X|
}
\hline
\textbf{Method} & \textbf{Memory Mechanisms} & \textbf{Benefits} \\
\hline

ECHO~\cite{echo}
& Hierarchical episodic memory
& Enhancing retrieval efficiency through structured memory \\
\hline

\textbf{HiMEM (Ours)}
& Hierarchical episodic memory + semantic memory consolidation + episodic memory pruning + joint memory utilization
& Reduced memory noise and redundancy, improved storage efficiency, and enhanced cross-task generalization \\
\hline

\end{tabularx}
\end{table*}
\begin{figure*}[t]
\centering
\includegraphics[width=0.8\textwidth]{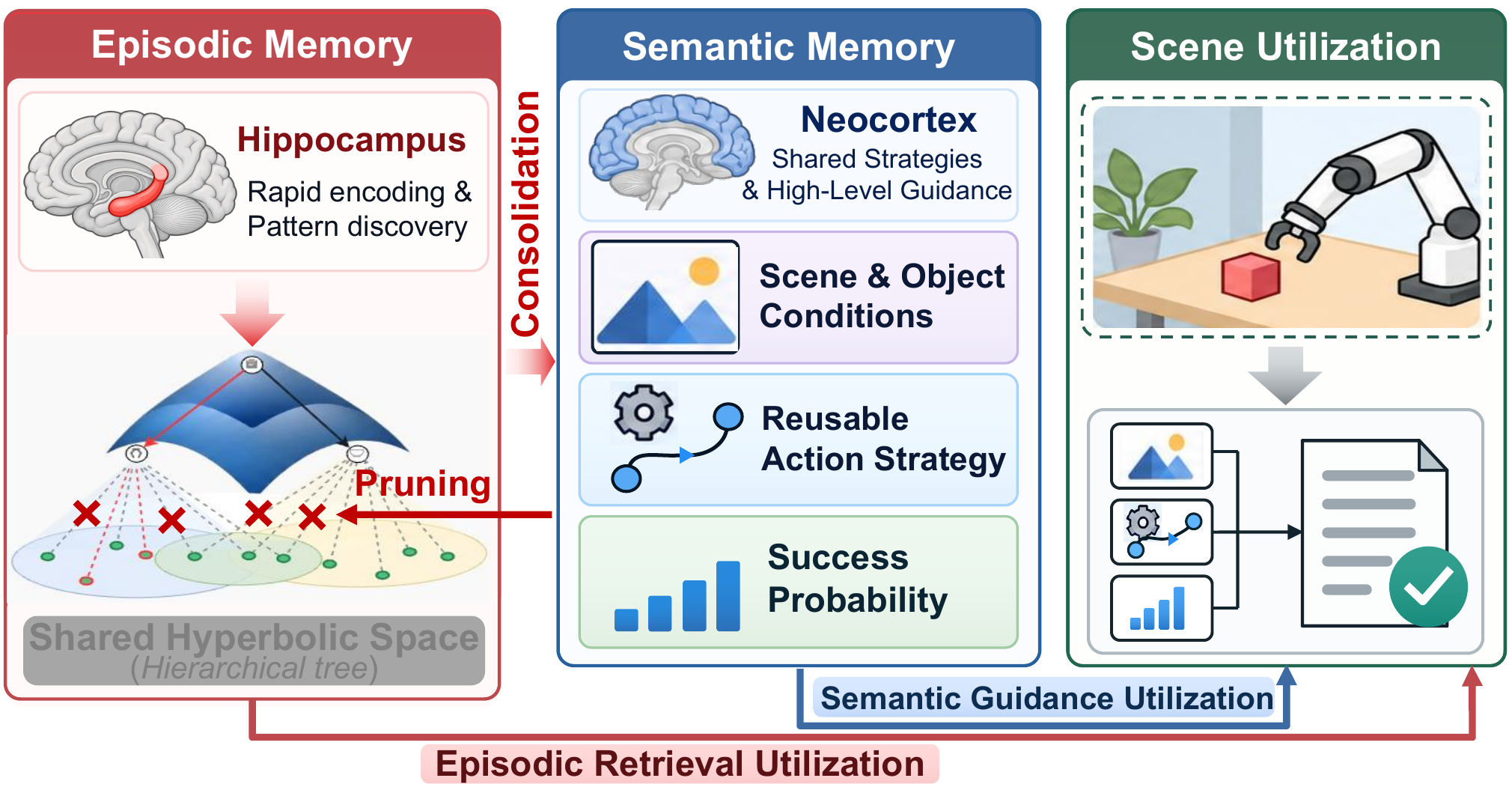}
\caption{\textbf{Overview of the episodic and semantic memory framework in HiMEM.} The hippocampus-inspired module organizes manipulation experiences into a shared hierarchical hyperbolic memory tree. The neocortex-inspired module consolidates experiences into compact semantic memories capturing scene and object conditions, reusable manipulation strategies, and success probabilities. Semantic consolidation guides the pruning of redundant episodic memories, while episodic retrieval and semantic guidance jointly support manipulation in complex 3D scenes.}

%
\label{fig:TREEvsFLAT}
\end{figure*}

This paper substantially extends our NeurIPS 2026 conference paper, ECHO \cite{echo}, to address long-horizon, complex spatial manipulation tasks in 3D environments. Table~\ref{tab:method_comparison} compares the memory mechanisms and benefits of the conference and journal versions. Building upon the original hierarchical episodic memory framework, this journal version further introduces semantic memory consolidation, episodic memory pruning, and joint utilization of episodic and semantic memories. The new contributions are summarized as follows:

\begin{enumerate}
    \item We propose a meta-semantic memory consolidation method inspired by the complementary mechanisms of the hippocampus and neocortex. Through cross-scenario statistical analysis, the method extracts shared patterns from manipulation experiences and organizes environmental conditions, object characteristics, manipulation strategies, and their execution success probabilities into compact semantic knowledge to enhance the transferability of experience across tasks.

    \item We propose an episodic memory pruning mechanism that removes large amounts of content-similar episodic memories after the corresponding generalizable semantic memories have been established, thereby improving the storage efficiency and capacity of long-term memory.

    \item We propose a task-conditioned memory composition and utilization mechanism that integrates episodic and meta-semantic memories. Based on the characteristics of the current scenario and manipulated objects, this mechanism combines historical experiences at different levels of abstraction and provides guidance for action generation through step-by-step task planning, thereby supporting experience reuse under novel task combinations and manipulation conditions.
\end{enumerate}

\begin{figure*}[t]
\centering
\includegraphics[width=\linewidth]{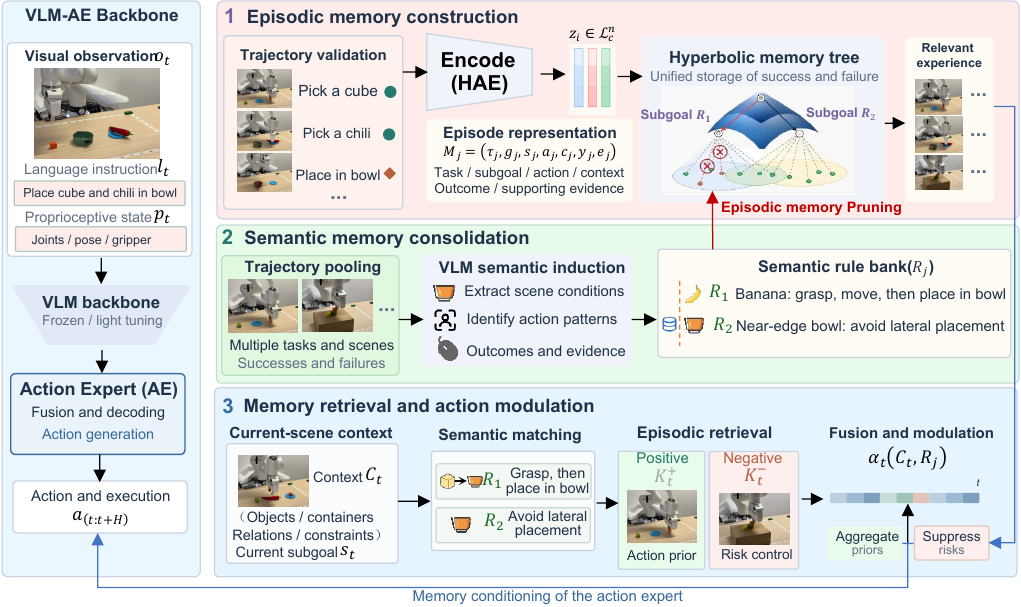}
\caption{\textbf{Overview of HiMEM for long-horizon spatial manipulation.} During offline memory formation, verified execution episodes are organized into a shared hierarchical hyperbolic memory, while a VLM consolidates cross-episode experiences into compact meta-semantic rules that capture scene conditions, manipulation strategies, and outcome-based success probabilities. During online execution, the VLA backbone provides hidden states and scene context for hierarchical episodic retrieval and semantic rule matching. The matched rules guide the weighting of successful episodic memories and suppress risky operations using failure evidence. The resulting semantic-conditioned memory is fused with short-term representations in Lorentz space, decoded into a residual vector, and injected into the VLA action-generation pathway to support long-horizon execution and cross-task generalization.}
\label{fig:framework}
\end{figure*}

%% file: sections/02_related.tex
\section{Related Work}
\label{sec:related_work}

\paragraph{Vision-Language-Action Models}
Vision-Language-Action (VLA) models map visual observations and natural-language instructions to robot actions. Existing approaches employ autoregressive action generation~\cite{rt2,openvla}, diffusion or flow-based policies~\cite{diffusion-policy,octo,pi0}, and efficient action heads~\cite{openvla-oft}. Dual-system architectures further separate high-level reasoning from low-level action generation~\cite{dpvla,robodual,fisvla}. While these architectures advance robotic manipulation, long-horizon tasks require retaining and reusing historical information beyond the current observation. This motivates external memory mechanisms that complement VLA policies with persistent experience, supporting long-horizon execution and knowledge reuse across tasks.

\paragraph{Memory for Long-Horizon Embodied Control}
Memory is essential for embodied control, as historical observations and experiences provide temporal context that may be absent from the current state~\cite{rt1,chen2021decision,raea,statler}. Recent studies explore different forms of memory representation and retrieval. MAP-VLA constructs stage-specific soft prompts from demonstrations and retrieves them through trajectory similarity~\cite{map-vla}. MemoryVLA integrates working memory with a persistent perceptual-cognitive memory bank, retrieving relevant representations and merging redundant entries~\cite{memoryvla}. MEM combines short-horizon video memory with long-horizon textual memory to capture events at different temporal and semantic scales~\cite{mem}. MemER selects task-relevant historical keyframes to support long-horizon robotic control~\cite{memer}. These methods demonstrate the importance of selecting, retaining, and reusing historical information. Beyond temporal dependencies within individual tasks, cross-task generalization further requires accumulating experiences across executions, organizing their outcomes, and extracting reusable knowledge from repeated interactions.

\paragraph{Hierarchical Episodic Memory Organization}
Episodic memory preserves specific experiences, including observations, task contexts, and action trajectories. As experiences accumulate, flat memory structures provide limited explicit organization of general-to-specific relationships, complicating memory navigation and reuse. Hierarchical memory representations have therefore attracted increasing attention. For instance, Embodied-RAG organizes spatial and semantic information into a hierarchical semantic forest, supporting retrieval at different levels of abstraction~\cite{embodied-rag}. Hyperbolic embeddings provide a continuous geometric representation of hierarchical relationships~\cite{rthe,poincare-emb}. In particular, the Lorentz model supports hierarchical representation learning~\cite{nickel2018learning}, while entailment cones model asymmetric containment relations~\cite{ganea2018hyperbolic}. Building on these foundations, HiMEM focuses on the hierarchical organization of manipulation experiences. It organizes task-, subgoal-, and action-level episodic memories within a shared continuous hyperbolic tree, supporting hierarchical retrieval and incremental memory insertion. Both successful and failed experiences are retained as episodic evidence, providing a structured basis for subsequent knowledge consolidation.

\paragraph{Episodic-to-Semantic Memory Consolidation}
Complementary learning systems theory distinguishes the rapid acquisition of specific experiences from the gradual integration of generalized knowledge across experiences~\cite{mcclelland1995there,kumaran2016learning}. This distinction motivates episodic-to-semantic memory consolidation, through which detailed experiences are transformed into reusable knowledge. Existing embodied memory methods have explored redundancy reduction and semantic abstraction. For example, MemoryVLA consolidates overlapping memory representations~\cite{memoryvla}, while MEM maintains abstract task-level information through long-horizon textual memory~\cite{mem}. However, accumulated episodic memories may contain extensive scene-specific, embodiment-related, and noisy details, making it difficult to extract common knowledge that generalizes across tasks. Effective transfer requires identifying the environmental and object conditions under which particular operations are applicable, together with their observed outcomes. Motivated by hippocampus-inspired consolidation, HiMEM employs video-language analysis to identify recurring conditions and operation patterns across recorded execution episodes. Verified success and failure outcomes provide statistical evidence for consolidating these patterns into compact, conditional semantic memories. The resulting knowledge directly conditions latent memory updates, guiding the integration of useful experiences and the suppression of unfavorable operation patterns. This connects hierarchical episodic memory with cross-execution knowledge consolidation and context-adaptive memory utilization for VLA models.

%% file: sections/03_method.tex
\section{Methodology}
\label{sec:method}


HiMEM is a long-term memory framework for VLA policies that integrates continuous hierarchical episodic memory with cross-episode meta-semantic consolidation. It segments executions into subgoals, encodes them in hyperbolic space, and organizes verified experiences into a general-to-specific hierarchy. Episodic memory supplies task-relevant successful manipulation priors; meta-semantic memory summarizes success and failure evidence into condition--operation rules that modulate prior aggregation and residual injection. Together, they condition action generation through lightweight backbone-specific projections. The overall architecture is illustrated in Fig.~\ref{fig:framework}.

\subsection{Episodic Memory Construction and Organization}
\label{subsec:memory_construction}

HiMEM builds upon ECHO's subgoal-level episodic memory representation, partitioning continuous executions into reusable manipulation segments. Qwen2-VL-7B-Instruct proposes task decompositions and subgoal boundaries. Each segment is represented as $M_i=(z_i,g_i^{\mathrm{task}},s_i,a_i,c_i,y_i,e_i)$, containing its hyperbolic embedding, global task, local subgoal, actions, scene conditions, execution outcome, and supporting evidence. Local outcomes are independently verified using simulator state predicates or human-verified real-robot records, avoiding incorrect attribution of whole-task failures to individual subgoals.

To organize these experiences, a hyperbolic autoencoder maps backbone hidden states $h_i$ into a Lorentz space with 512 spatial dimensions, yielding $z_i=E_\phi(h_i)$. Feature reconstruction and task--subgoal entailment constraints support hierarchical representations, organizing episodic memories into a general-to-specific tree. Successful segments provide retrievable manipulation priors, while both successful and failed experiences preserve contextual and outcome evidence. This episodic memory structure serves as the foundation for the subsequent semantic consolidation, redundancy pruning, and memory composition mechanisms.

\subsection{Cross-Episode Meta-Semantic Consolidation}
\label{subsec:semantic_consolidation}


Episodic memory preserves individual experiences; meta-semantic memory abstracts when operations are applicable~\cite{kumaran2016learning}. Offline consolidation selects related operations with different conditions or outcomes across branches. Comparing their videos and states, a VLM proposes manipulation-relevant conditions $C_j$ and operations $A_j$, forming rules with conditional success estimates: 
\begin{equation}
 R_j=(C_j,A_j,\widehat P_j),
 \qquad \widehat P_j\approx P(\mathrm{success}\mid C_j,A_j).
 \label{eq:semantic_rule}
\end{equation}
Each rule retains source evidence and a version. Verified outcomes supply success and failure counts $n_j^+,n_j^-$, with each episode counted at most once per rule. Duplicates are removed; conflicts and unresolved outcomes are withheld. A Beta prior with $a_0=b_0=1$ gives 
\begin{equation}
 \widehat P_j=\frac{n_j^++a_0}{n_j^++n_j^-+a_0+b_0}.
 \label{eq:rule_probability}
\end{equation}
Identical rules merge; specific conditions link to general ones for backoff. Revised conditions or labels trigger recounting. The resulting store supplies evidence-based applicability estimates for retrieved operations.

Qwen2-VL-7B-Instruct provides task-decomposition and subgoal-boundary proposals. Each episodic record stores the episode identifier, start and end times, attempted operation, context predicates, local outcome, outcome-test identifier, and references to the corresponding video and state evidence. Task-level success is recorded separately from local subgoal outcomes, and records with unresolved attribution remain available for review. In simulation, local outcomes are verified against MuJoCo states and BDDL predicates, whereas physical outcomes rely on human-verified execution records. Subgoal-boundary detection and local outcome attribution are handled separately to distinguish temporal segmentation from execution success.

\subsection{Semantic-Conditioned Episodic Pruning}
\label{subsec:semantic_pruning}

Once a meta-semantic rule is consolidated, HiMEM applies semantic-conditioned diversity-preserving pruning (SCDP) to remove redundant episodic vectors. For a rule $R_j$ with verified source evidence, existing rule--episode links identify its provenance set $\mathcal E_j=\{M_i:i\in\operatorname{Prov}(R_j)\}$. Each record receives a condition signature $b_i=\operatorname{sig}(c_i,a_i,y_i)$ derived from normalized scene predicates, discretized continuous attributes, attempted operations, and verified local outcomes. Only records whose provenance and operation match $R_j$ and whose context satisfies $C_j$ are eligible for pruning. Partitioning $\mathcal E_j$ into $\mathcal E_{j,b}=\{M_i\in\mathcal E_j:b_i=b\}$ preserves separate representatives for distinct conditions and outcomes.

For each condition group, SCDP selects representative episodes in the existing Lorentz space to minimize the maximum distance to the retained subset. The protected set $\mathcal U_{j,b}$ contains distinctive failures, condition-boundary cases, and other non-substitutable evidence. Farthest-first traversal starts from protected records (or a verified exemplar when the set is empty) and adds the farthest record until the target budget $K_{j,b}$ is reached or the maximum coverage distance is within $\epsilon$. The budget is relaxed when the protected set exceeds it. The selection objective for a feasible budget is
\begin{equation}
\begin{aligned}
\min_{\mathcal S\subseteq\mathcal E_{j,b}}\quad&
\max_{M_i\in\mathcal E_{j,b}}\min_{M_k\in\mathcal S}
 d_{\mathcal L}(z_i,z_k),\\
\text{s.t.}\quad&1\leq|\mathcal S|\leq K_{j,b},\quad
\mathcal U_{j,b}\subseteq\mathcal S.
\end{aligned}
\label{eq:scdp_selection}
\end{equation}

A source record is pruned only when it is covered by the consolidated rule, represented by a nearby record with the same condition signature, and dispensable to every other linked rule. Specifically, $d_{\mathcal L}(z_i,\mathcal S)=\min_{M_k\in\mathcal S}d_{\mathcal L}(z_i,z_k)$; $\operatorname{Cov}$ verifies provenance and rule applicability, $\operatorname{Unique}$ protects rare failures, condition boundaries, and distinctive state transitions, and $\operatorname{Safe}_{\mathrm{all}}$ requires that all other linked rules retain a same-signature representative or protected evidence. The deletion criterion is
\begin{equation}
\begin{aligned}
\operatorname{Prune}(M_i)=1\iff{}&
\operatorname{Cov}(M_i,R_j)\land\neg\operatorname{Unique}(M_i)\\
&\land d_{\mathcal L}(z_i,\mathcal S_{j,b_i})\leq\epsilon\\
&\land\operatorname{Safe}_{\mathrm{all}}(M_i).
\end{aligned}
\label{eq:scdp_pruning}
\end{equation}

Pruning preserves the original verified success/failure counts and conditional statistics of each rule, so representative selection does not alter $\widehat P_j$. A compact provenance ledger retains episode identifiers, conditions, operations, outcomes, and references to original evidence for subsequent rule revision and recounting. Eligible leaf records are removed from episodic storage and vector indices, while parent-node summaries and rule--source links are updated without directly deleting structural cones. The pruned tree and semantic rules are committed in a joint versioned snapshot for consistent online retrieval.

\subsection{Hierarchical Retrieval and Semantic Conditioning}
\label{subsec:retrieval}


HiMEM first retrieves relevant experiences geometrically, then assesses their operations using rules. For the $\pi_0$ backbone, the query state $h_t$ is obtained from the final VLM prefix token. The query $q_t=E_\phi(h_t)$ searches with beam width $k=3$, expanding containing cones and ranking leaves by Lorentz similarity:
\begin{equation}
 \varphi(q_t,z_\mu)\leq\omega(z_\mu),
\end{equation}
 
\begin{equation}
\begin{aligned}
 d_{\mathcal L}(z,q)&=\frac{\operatorname{acosh}(-c\langle z,q\rangle_{\mathcal L})}{\sqrt c},\\
 \operatorname{sim}(z,q)&=\frac{1}{1+d_{\mathcal L}(z,q)}.
\end{aligned}
\label{eq:retrieval_similarity}
\end{equation}
 Successful candidates $\mathcal K_t^+$ supply priors; failed candidates reference outcome statistics without recounting. Scene conditions $C_t$, cached at task onset, subgoal transitions, or relevant scene changes, select the most specific supported matching rule, with backoff to general conditions. Unknown or unsatisfied required predicates exclude rules. Success $\widehat P_i$, support $n_i$, and match $m_i\in[0,1]$ determine 
\begin{equation}
 u_i=m_i\frac{n_i}{n_i+\kappa}(2\widehat P_i-1),
 \label{eq:semantic_reliability}
\end{equation}
 
\begin{equation}
 g_i=\begin{cases}
 0,&n_i\geq n_{\min}\ \text{and}\ \widehat P_i<\tau_{\mathrm{risk}},\\
 \exp(\lambda_{\mathrm{sem}}u_i),&\text{otherwise}.
 \end{cases}
 \label{eq:semantic_coefficient}
\end{equation}
 Here $\kappa$ shrinks limited support, $\lambda_{\mathrm{sem}}$ controls modulation, and $n_{\min},\tau_{\mathrm{risk}}$ govern blocking. The coefficient strengthens, attenuates, or blocks candidates. Same-key local statistics can substitute for unavailable rules; absent evidence gives $u_i=0,g_i=1$.

\subsection{Geometric Memory Composition}
\label{subsec:synthesis}


A retrieved experience may cover only part of the current operation. To compose related priors, HiMEM optionally interpolates between compatible, unblocked successful memories along a Lorentz geodesic: 
\begin{equation}
 z_{\mathrm{syn}}=\operatorname{Exp}_{z_i}
 \!\left(\rho\operatorname{Log}_{z_i}(z_j)\right).
 \label{eq:memory_synthesis}
\end{equation}
 Normalized query similarities determine $\rho$. Each synthesized candidate $s$, with $z_s=z_{\mathrm{syn}}$, retains its sources $(i,j)$, inherits the conservative coefficient $g_s=\min(g_i,g_j)$, and joins $\widetilde{\mathcal K}_t^+=\mathcal K_t^+\cup\mathcal S_t$ for subsequent aggregation and gating. Without synthesis, $\mathcal S_t=\emptyset$. Synthetic vectors serve current inference only and enter neither the tree nor rule counts; permanent memory accepts verified executions only.

\subsection{Memory Fusion and Policy Injection}
\label{subsec:integration}


To form an action prior, HiMEM applies softmax over $\widetilde{\mathcal K}_t^+$, including synthetic candidates, and reweights by $g_i$: 
\begin{equation}
 \pi_i^{\mathrm{geo}}=
 \frac{\exp(\operatorname{sim}(z_i,q_t))}
 {\sum_{\ell\in\widetilde{\mathcal K}_t^+}\exp(\operatorname{sim}(z_\ell,q_t))},
\end{equation}
 
\begin{equation}
 \begin{gathered}
 \pi_i^{\mathrm{sem}}=
 \frac{\pi_i^{\mathrm{geo}}g_i}
 {\sum_{\ell\in\widetilde{\mathcal K}_t^+}\pi_\ell^{\mathrm{geo}}g_\ell},
 \\
 z_{\mathrm{mem}}^{\mathrm{sem},+}
 =\Pi_{\mathcal L}\!\left(\sum_{i\in\widetilde{\mathcal K}_t^+}\pi_i^{\mathrm{sem}}z_i\right).
 \end{gathered}
 \label{eq:semantic_weights}
\end{equation}
 Projection $\Pi_{\mathcal L}(v)=v/\sqrt{-c\langle v,v\rangle_{\mathcal L}}$ normalizes future-directed time-like vectors. The highest unblocked similarity $s_t^+$ determines geometric gating, while bounded risk attenuates injection strength: 
\begin{equation}
\begin{aligned}
 \beta_t&=\sigma\!\left(5(s_t^+-0.5)\right),\\
 g_{\mathrm{risk},t}&=\sum_{i\in\widetilde{\mathcal K}_t^+}
       \pi_i^{\mathrm{geo}}\min(1,g_i),\\
 \alpha_t&=\alpha_0\beta_tg_{\mathrm{risk},t},\qquad \alpha_0=0.03.
\end{aligned}
\label{eq:injection_strength}
\end{equation}
 HAE-encoded, pooled short-term states $z_{\mathrm{short}}$ combine with the long-term prior. Decoding produces a correction relative to $h_t$: 
\begin{equation}
\begin{aligned}
 z_{\mathrm{ctx}}&=\Pi_{\mathcal L}\!\left((1-\beta_t)z_{\mathrm{short}}
                 +\beta_t z_{\mathrm{mem}}^{\mathrm{sem},+}\right),\\
 \Delta_t&=D_\psi(z_{\mathrm{ctx}})-h_t.
\end{aligned}
\label{eq:conditioned_context}
\end{equation}
 For $\pi_0$, the residual is projected to the Action Expert width $d_A$ and broadcast identically to all $N_{\mathrm{suf}}$ suffix tokens: 
\begin{equation}
\begin{gathered}
 r_t=\operatorname{proj}(\alpha_t\Delta_t)\in\mathbb R^{d_A},\\
 E'_{\mathrm{suffix},\ell}=E_{\mathrm{suffix},\ell}+r_t,
 \quad \ell=1,\ldots,N_{\mathrm{suf}}.
\end{gathered}
 \label{eq:policy_injection}
\end{equation}
Other backbones use corresponding interfaces. Empty or fully blocked candidates, or invalid normalization, disable injection. Successful experiences thus supply correction directions, while outcome statistics regulate their contribution and strength; failed vectors never supply action priors.

During online execution, the policy reads the committed snapshot and uses cached scene conditions for geometric retrieval and rule matching. Failure retrieval accesses existing evidence summaries without updating episode counts online; only eligible successful memories contribute action-prior vectors. When no applicable semantic rule is available, the semantic coefficient defaults to one, whereas an empty or fully blocked positive candidate set forces the residual strength to zero. The semantic coefficient is applied once during candidate aggregation, while the separate bounded risk term controls the final residual amplitude.

%% file: sections/04_experiments.tex
\section{Experiments}
\label{sec:experiments}

\begin{table*}[t]
  \caption{
    Success rates (\%) on LIBERO, LIBERO-Plus.
    LIBERO \textbf{Avg.} averages the four standard suites.
    Shaded rows denote matched +HiMEM comparisons.
    \textsuperscript{ZS} denotes zero-shot evaluation.
    A dash (--) indicates that the method was not evaluated on the corresponding benchmark.
  }
  \label{tab:main_results}

  \centering
  \footnotesize
  \renewcommand{\arraystretch}{1.10}
  \setlength{\tabcolsep}{2.6pt}

  \begin{tabularx}{\textwidth}{
    @{}
    >{\raggedright\arraybackslash}p{3.5cm}
    >{\centering\arraybackslash}X
    >{\centering\arraybackslash}X
    >{\centering\arraybackslash}X
    >{\centering\arraybackslash}X
    >{\centering\arraybackslash}X
    >{\centering\arraybackslash}X
    @{}
  }
    \toprule
    \multirow{2}{*}{Method}
      & \multicolumn{4}{c}{LIBERO}
      & \multirow{2}{*}{Avg.}
      & \multirow{2}{*}{\shortstack{LIBERO-\\Plus}}
    \\
    \cmidrule(lr){2-5}
      & Spatial & Object & Goal & Long & & \\
    \midrule

    \multicolumn{7}{@{}l}{\textbf{\textit{Reported baselines}}} \\[-2pt]

    Octo~\cite{octo}
      & 78.9 & 85.7 & 84.6 & 51.1 & 75.08 & --\\

    OpenVLA~\cite{openvla}
      & 84.7 & 88.4 & 79.2 & 53.7 & 76.50 & 15.6\\

    CronusVLA~\cite{Cronusvla}
      & 97.3 & 99.6 & 96.9 & 94.0 & 97.00 & --\\

    MAP-VLA~\cite{map-vla}
      & 96.3 & 98.4 & 95.4 & 83.4 & 93.38 & --\\

    MemoryVLA~\cite{memoryvla}
      & 98.4 & 98.4 & 96.4 & 93.4 & 96.65 & 70.2\\

    \midrule

    \multicolumn{7}{@{}l}{\textbf{\textit{HiMEM across VLA backbones}}} \\[-2pt]

    $\pi_0$~\cite{pi0}
      & 97.5 & 97.0 & 92.3 & 80.7 & 91.88 & 54.2$^{\mathrm{ZS}}$\\

    $\pi_0$ + ECHO\cite{echo}
      & 98.3 & 98.8 & 98.6 & 93.5 & 97.30 & 56.9$^{\mathrm{ZS}}$\\

    \rowcolor{gray!12}
    $\pi_0$ + \textbf{HiMEM}
      & \textbf{99.8}
      & \textbf{98.8}
      & \textbf{98.9}
      & \textbf{95.3}
      & \textbf{98.20}
      & \textbf{59.6}$^{\mathrm{ZS}}$\\

    \addlinespace[1.5pt]
    \cmidrule[0.3pt](lr){1-7}
    \addlinespace[1.5pt]

    $\pi_{0.5}$~\cite{pi05}
      & 99.5 & 95.5 & 93.6 & 91.6 & 95.05 & 85.7\\

    $\pi_{0.5}$ + ECHO\cite{echo}
      & 99.5 & 97.5 & 98.5 & 94.6 & 97.53 & 92.7\\

    \rowcolor{gray!12}
    $\pi_{0.5}$ + \textbf{HiMEM}
      & \textbf{99.8}
      & \textbf{99.8}
      & \textbf{98.7}
      & \textbf{96.5}
      & \textbf{98.70}
      & 95.8\\

    \addlinespace[1.5pt]
    \cmidrule[0.3pt](lr){1-7}
    \addlinespace[1.5pt]

    OpenVLA-OFT~\cite{openvla-oft}
      & 97.6 & 98.4 & 97.9 & 93.3 & 96.80 & 69.6\\

    OpenVLA-OFT + ECHO\cite{echo}
      & 98.0 & 99.2 & 99.6 & 96.7 & 98.38 & 71.9\\

    \rowcolor{gray!12}
    OpenVLA-OFT + \textbf{HiMEM}
      & \textbf{98.8}
      & \textbf{99.6}
      & \textbf{99.7}
      & \textbf{97.7}
      & \textbf{98.95}
      & \textbf{74.2}\\

    \bottomrule
  \end{tabularx}
\end{table*}

\begin{table}[t]
\centering
\renewcommand{\arraystretch}{1.12}
\footnotesize
\caption{Physical manipulation success rates (\%). Averages are reported separately for basic and composite tasks.}
\label{tab:robot_comparison}
\setlength{\tabcolsep}{3pt}
\begin{tabularx}{\linewidth}{@{}>{\raggedright\arraybackslash}Xccc@{}}
\toprule
Task & Backbone & ECHO & HiMEM \\
\midrule
Place & 75.0 & 90.0 & 90.0 \\
Stack & 45.0 & 55.0 & 70.0 \\
Insert & 55.0 & 65.0 & 75.0 \\
\textbf{Basic-task average} & \textbf{58.3} & \textbf{70.0} & \textbf{78.3} \\
\midrule
G1 & 40.0 & 50.0 & 65.0 \\
G2 & 45.0 & 55.0 & 60.0 \\
G3 & 45.0 & 60.0 & 70.0 \\
G4 & 35.0 & 40.0 & 55.0 \\
G5 & 20.0 & 30.0 & 55.0 \\
\textbf{Composite-task average} & \textbf{37.0} & \textbf{47.0} & \textbf{61.0} \\
\bottomrule
\end{tabularx}
\end{table}

Our experiments address four questions: (1) Does HiMEM consistently improve different VLA backbones on standard robotic manipulation benchmarks? (2) Can it reuse experience when target-suite, target-task, or complete composite trajectories are absent from memory? (3) How do hierarchical episodic memory and cross-episode meta-semantic consolidation jointly influence the policy, and what does each component contribute? (4) Does the system maintain efficient retrieval and practical deployment costs as memory grows?

\begin{figure*}[!tbp]
  \centering
  \includegraphics[width=\textwidth]{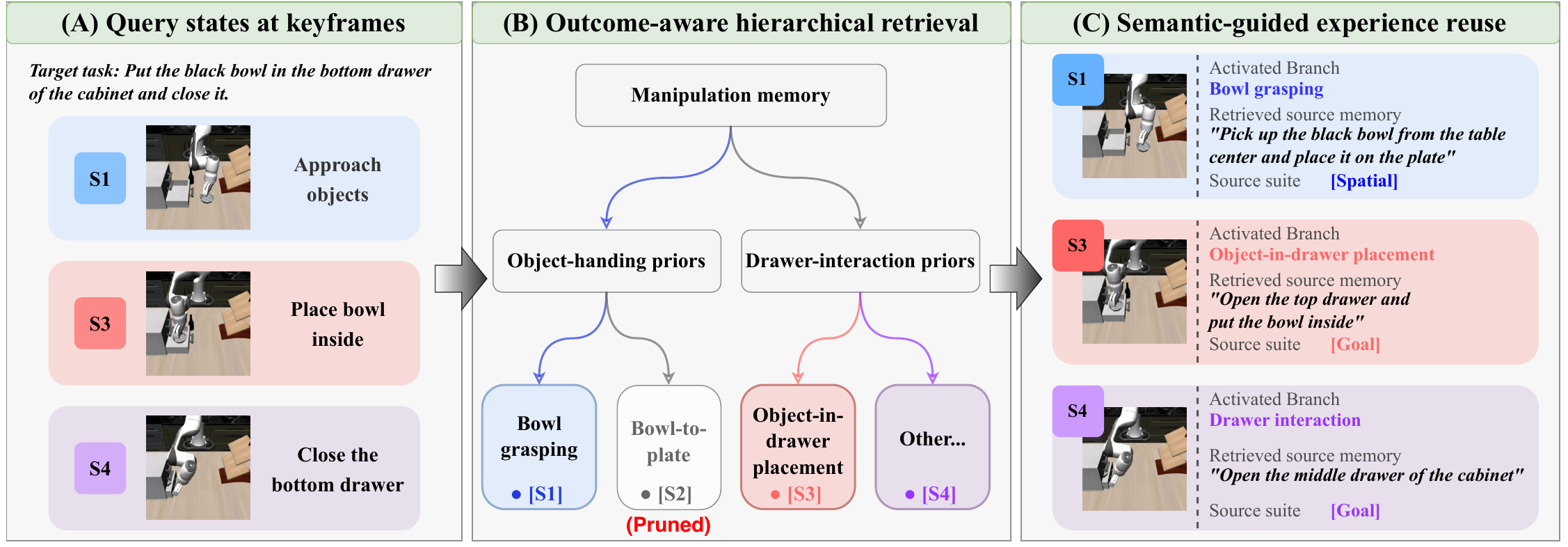}
    \caption{Hierarchical memory retrieval and semantic-guided cross-suite compositional experience reuse. (A) Representative query states at three key execution steps (S1, S3, and S4) of the target task. (B) Coarse-to-fine traversal of the hierarchical manipulation memory, illustrating activated retrieval paths and pruned irrelevant branches. (C) Historical experiences retrieved from the Spatial and Object suites provide reusable manipulation priors for grasping, in-drawer placement, and drawer closure, enabling compositional experience transfer across subgoals.}

  \label{fig:geometric_retrieval_example}
\end{figure*}

\subsection{Experimental Setup}
\label{subsec:experiment_setup}

We use the four LIBERO suites~\cite{libero} as the primary simulation benchmark, LIBERO-Plus~\cite{libero-plus} for robustness, and a Franka Emika Panda for transfer and physical evaluation. Main results compare matched backbone/HiMEM pairs for $\pi_0$~\cite{pi0}, $\pi_{0.5}$~\cite{pi05}, and OpenVLA-OFT~\cite{openvla-oft}. Component contributions are assessed by ablation. Successful segments supply action priors, while verified successful and failed outcomes support rule statistics. Memory construction, validation, and testing use disjoint executions. The policy and memory remain fixed during evaluation; paired groups share checkpoints, initial states, seeds, and action horizons. 

For LIBERO-Spatial, Object, Goal, and Long, the policy receives 224$\times$224 third-person and wrist RGB observations, the language instruction, and 8-D proprioception. Five predicted actions are executed per replan. Maximum episode horizons are 220, 280, 300, and 520 steps for Spatial, Object, Goal, and Long, respectively, with a common 10-step stabilization convention and no retry after failure.

Simulation memory is initialized from ten successful demonstrations per task, while separately collected failure evidence comes from the same task distribution and remains disjoint from evaluation. The $\pi_0$ comparison retains the same 30k-step policy checkpoint across methods. Checkpoint identifiers, benchmark revisions, and corresponding run configurations are recorded in the experiment manifest.

Transition-detection supervision is constructed from official HDF5 MuJoCo states and BDDL predicates: frames satisfying a relevant predicate are positive samples, while earlier frames from the same demonstration provide transition negatives. These transition negatives are distinct from failed manipulation experiences. Qwen2-VL-7B-Instruct is adapted with LoRA using structured prompts and JSON outputs, with trajectory-disjoint training and validation splits. Boundary detection, local outcome attribution, and cross-video rule induction are evaluated separately, with each prediction checked against its corresponding state or human-verified record.

\subsection{Simulation Performance and Robustness}
\label{subsec:main_results}

Table~\ref{tab:main_results} summarizes the four LIBERO suites and LIBERO-Plus. Published methods provide cross-paper references; matched backbone/HiMEM comparisons assess the complete method. Long emphasizes multi-step experience reuse, while LIBERO-Plus tests perturbation robustness. Each LIBERO task uses 50 episodes under seeds 7, 123, and 2024; the aggregate score averages the four suite success rates.

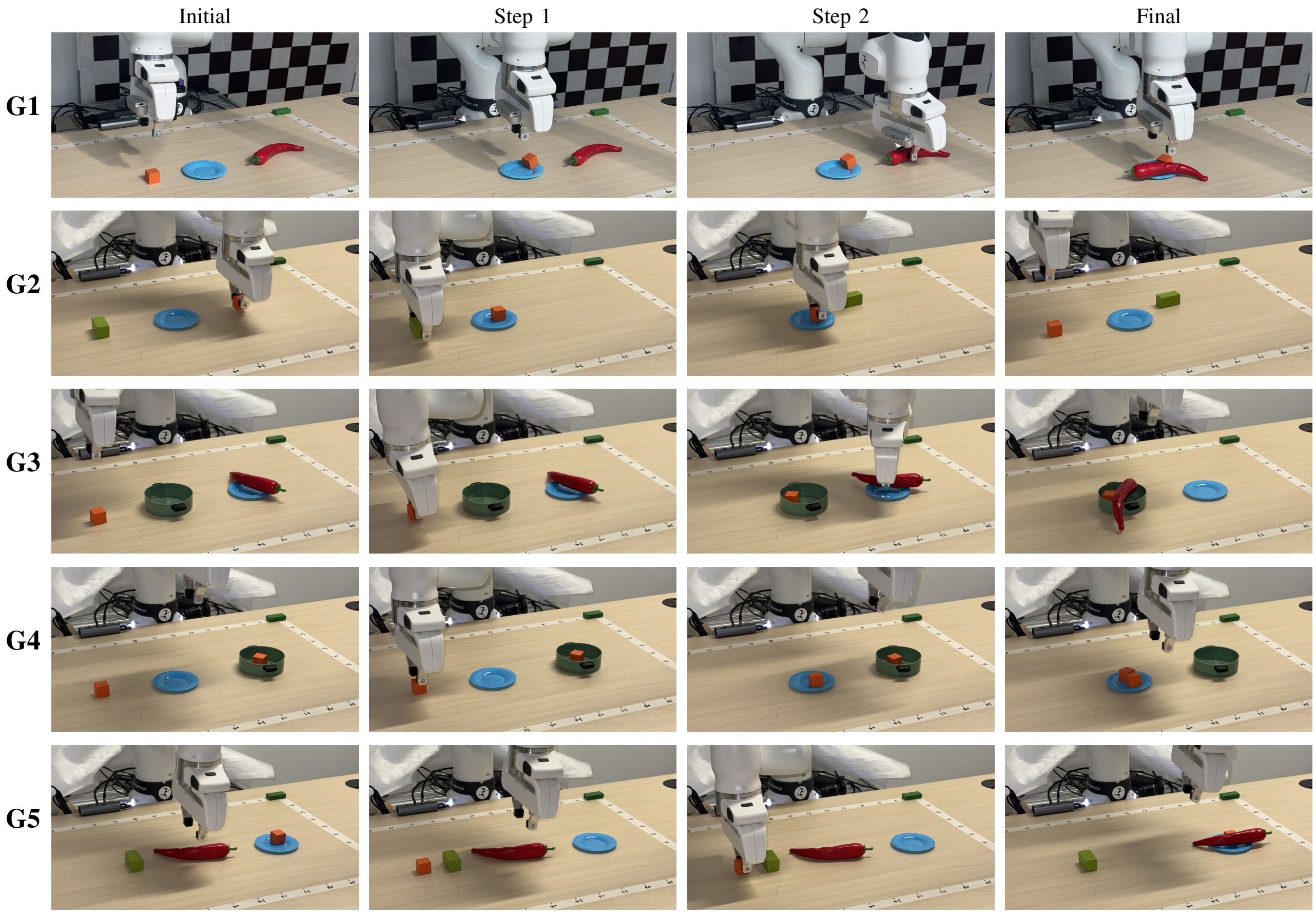
\begin{figure*}[t]
  \centering
  \begingroup
  \setlength{\tabcolsep}{2pt}
  \renewcommand{\arraystretch}{1.0}

  \newlength{\realworldW}
  \newlength{\realworldH}
  \setlength{\realworldW}{%
    \dimexpr0.205\textwidth+2\fboxsep+2\fboxrule\relax}
  \setlength{\realworldH}{%
    \dimexpr1.88cm+2\fboxsep+2\fboxrule\relax}

  \newcommand{\realworldcrop}[2]{%
    \begin{tikzpicture}[baseline=(current bounding box.center)]
      \path[use as bounding box]
        (0,0) rectangle (\realworldW,\realworldH);
      \begin{scope}
        \clip (0,0) rectangle (\realworldW,\realworldH);
        \node[inner sep=0,outer sep=0]
          at (\dimexpr\realworldW/2\relax,
              \dimexpr\realworldH/2\relax) {%
          \includegraphics[
            height=\realworldW,
            angle=-90
          ]{figures/realworld/#1-#2.jpg}%
        };
      \end{scope}
    \end{tikzpicture}%
  }

  \begin{tabular}{@{}c@{\hspace{4pt}}cccc@{}}
    & \footnotesize Initial
    & \footnotesize Step 1
    & \footnotesize Step 2
    & \footnotesize Final \\

    \textbf{G1}
    & \realworldcrop{1}{1}
    & \realworldcrop{1}{2}
    & \realworldcrop{1}{3}
    & \realworldcrop{1}{4}
    \\\noalign{\vskip 5pt}

    \textbf{G2}
    & \realworldcrop{2}{1}
    & \realworldcrop{2}{2}
    & \realworldcrop{2}{3}
    & \realworldcrop{2}{4}
    \\\noalign{\vskip 5pt}

    \textbf{G3}
    & \realworldcrop{3}{1}
    & \realworldcrop{3}{2}
    & \realworldcrop{3}{3}
    & \realworldcrop{3}{4}
    \\\noalign{\vskip 5pt}

    \textbf{G4}
    & \realworldcrop{4}{1}
    & \realworldcrop{4}{2}
    & \realworldcrop{4}{3}
    & \realworldcrop{4}{4}
    \\\noalign{\vskip 5pt}

    \textbf{G5}
    & \realworldcrop{5}{1}
    & \realworldcrop{5}{2}
    & \realworldcrop{5}{3}
    & \realworldcrop{5}{4}
  \end{tabular}
  \endgroup

  \caption{
    Representative real-robot rollouts for five held-out compositional tasks (G1--G5). Each row shows four temporally ordered frames from one execution.
  }
  \label{fig:qualitative_retrieval}
\end{figure*}

\subsection{Held-Out Experience Transfer}
\label{sec:generalization}

Suite-, task-, and composite-trajectory holdouts test whether HiMEM reuses knowledge without direct target experience. For LIBERO-Long, memory uses only Spatial, Object, and Goal experience, excluding all Long trajectories. Target executions are excluded before both episodic storage and rule consolidation to prevent indirect exposure through semantic evidence. 
Both memories remain frozen throughout each complete evaluation. Although the general deployment procedure permits validated write-back between tasks, controlled target-experience holdout experiments disable this mechanism. Current observations may refresh cached scene-condition descriptions but cannot modify stored episodic memories or rule statistics. Where available, same-target memory provides a reference for distinguishing transfer under target-experience holdout from reuse of same-task experience.


\subsection{Qualitative Retrieval and Rule Analysis}
\label{subsec:qualitative}

Figure~\ref{fig:geometric_retrieval_example} presents an example of cross-suite compositional retrieval. The target instruction asks the policy to place a black bowl in the bottom drawer of a cabinet and close it. At three representative keyframes, the query activates memory branches corresponding to bowl grasping (S1), object-in-drawer placement (S3), and drawer closure (S4), retrieving relevant historical segments from LIBERO-Spatial and LIBERO-Object. Hierarchical retrieval identifies manipulation experiences relevant to the current subgoals while pruning irrelevant branches, such as bowl-to-plate placement. The retrieved experiences provide reusable manipulation priors for grasping, in-drawer placement, and drawer closure without replaying the complete target trajectory. Branch names are post-hoc summaries added for visualization. This example primarily illustrates hierarchical episodic retrieval and cross-suite compositional reuse, while the specific effects of semantic-rule matching and residual modulation require further analysis using the corresponding rule and execution records.


\subsection{Real-World Manipulation}
\label{subsec:real_world}
\label{subsec:real_world_extension}

We evaluate on a Franka Emika Panda using \emph{Place Banana in Bowl}, \emph{Stack Blocks}, and \emph{Insert Circle into Base}, covering semantic goal grounding, sequential manipulation, and precision-sensitive insertion. Each task uses 20 randomized initial configurations for matched backbone/episodic-only ablation/HiMEM comparisons with the same policy checkpoint. Each task uses 20 matched initial configurations, each evaluated with the backbone, ECHO, and HiMEM, yielding 60 rollouts per task. The final results record the actual completed trial counts and any deviations from the planned budget. Table~\ref{tab:robot_comparison} reports successes, trial counts, and mean success rates. Five held-out compositions, G1--G5, cover multi-step placement, sorting, stacking, and loaded-object transport, each with 20 matched configurations. Memory is constructed from 10 successful teleoperated demonstrations per atomic skill under standard layouts and separately collected, verified failed attempts. Successful segments supply manipulation priors; both outcomes support conditional rules. Complete composite experience and evaluation layouts are excluded from both memory sources. Evaluation permits no retry, online policy adaptation, or memory update. Execution sequences relate object and scene conditions to rule matches, examining experience reuse under physical variation.
The physical platform consists of a Franka Emika Panda arm, a parallel gripper, and a static third-person RGB camera. RGB observations are padded and resized to $224\times224$, and the policy server receives synchronized proprioceptive and gripper states through a WebSocket interface. The control-loop target is 15 Hz.

The five composite tasks evaluate compositional generalization to multi-step manipulation by recombining basic manipulation skills. The tasks involve colored blocks, bowls, trays, and a red chili pepper, with predefined initial layouts and multi-step execution sequences. \textbf{G1} sequentially places an orange block and a red chili pepper onto a blue tray; \textbf{G2} swaps the positions of an orange block and a green block through three consecutive transfers, using the blue tray as temporary storage; \textbf{G3} first places an orange block into a green bowl, then transfers a red chili pepper from the blue tray into the same bowl; \textbf{G4} transfers two orange blocks, initially located on the table and inside the green bowl, respectively, onto the blue tray, requiring object placement across containers; and \textbf{G5} temporarily relocates an orange block from the blue tray to the table, then rearranges the orange block and red chili pepper so that both are placed on the tray. By recombining sequential placement, temporary storage, position swapping, cross-container transfer, and space clearing, these tasks assess the transfer of previously acquired manipulation experiences to unseen multi-step task sequences.

For each atomic skill, ten successful teleoperated demonstrations covering different support conditions are used to construct shared episodic memory. ECHO and HiMEM utilize the same original episodic experiences, while HiMEM further consolidates generalizable manipulation patterns from these experiences. Complete trajectories, composite instructions, and evaluation layouts of the target tasks are excluded from memory to assess generalization to unseen manipulation conditions and task compositions.

\subsection{Ablation Study}
\label{sec:ablation}

Table~\ref{tab:ablation} presents ablation results on LIBERO-Long. We first compare short-term buffers, flat Euclidean memory, hyperbolic memory, and cone-tree retrieval to assess memory organization and retrieval mechanisms. Hyperbolic memory and cone-tree retrieval achieve success rates of 91.11\% and 92.04\%, respectively, compared with 80.70\% for the vanilla backbone. Removing virtual synthesis, failure evidence, or cross-episode consolidation reduces the success rate from 95.30\% to 92.69\%, 89.33\%, and 93.75\%, respectively, demonstrating their contributions. Compared with full ECHO (93.48\%), HiMEM further improves success by 1.82 percentage points, supporting the effectiveness of combining episodic and meta-semantic memories for long-horizon manipulation.

Each evaluation records the policy and HAE checkpoints, random seed, actual episode count, and source/evaluation split. Outcome accounting distinguishes successful, failed, and unresolved segments; semantic analyses additionally record rule support, rule coverage, matched-rule counts, and residual strength. Aggregate metrics follow the corresponding benchmark definitions, and hyperparameters are selected using validation data only.

\begin{table*}[t]
  \centering
  \caption{Ablations on LIBERO-Long. Geometric controls assess memory organization and retrieval; three semantic controls assess outcome evidence, cross-episode consolidation, and online conditioning.}
  \label{tab:ablation}
\label{tab:mechanism_ablation}
  \footnotesize
  \renewcommand{\arraystretch}{1.10}
  \setlength{\tabcolsep}{4.5pt}

  \begin{tabularx}{0.96\textwidth}{
    @{}
    >{\raggedright\arraybackslash}p{4.45cm}
    >{\raggedright\arraybackslash}X
    >{\centering\arraybackslash}p{1.55cm}
    @{}
  }
    \toprule
    Model Variant & Memory / Retrieval Configuration & Success (\%) \\
    \midrule
    Vanilla $\pi_0$ & -- & \textsc{80.70} \\
    $\pi_0$ + Short-term Buffer & Euclidean within-episode buffer & \textsc{88.81} \\
    $\pi_0$ + Flat Memory & Flat Euclidean $k$NN retrieval & \textsc{83.25} \\
    $\pi_0$ + Hyperbolic Memory & Hyperbolic distance-based $k$NN & \textsc{91.11} \\
    $\pi_0$ + Cone Tree & Entailment-aware cone-tree retrieval & \textsc{92.04} \\
    $\pi_0$ + \textit{HiMEM} w/o Virtual Synthesis & Cone tree + consolidation & \textsc{92.69} \\
    \midrule
    HiMEM w/o failure evidence & Successful evidence only; rule pipeline retained & \textsc{89.33} \\
    HiMEM w/o cross-episode consolidation & Local condition--operation statistics & \textsc{93.75} \\
    Full ECHO & -- & \textsc{93.48} \\
    \rowcolor{gray!15}
    \textbf{Full HiMEM} & \textbf{Episodic + meta-semantic memory} & \textbf{95.30} \\
    \bottomrule
  \end{tabularx}
\end{table*}

\subsection{Cross-Suite and Compositional Generalization}
\label{subsec:cross_suite_generalization}

We evaluate generalization through LIBERO cross-suite transfer and five real-robot compositional tasks, excluding target-suite or complete composite-task episodic memories. As shown in Table~\ref{tab:cross_suite_generalization}, ECHO improves LIBERO cross-suite success from 80.70\% to 89.31\%. On real-robot compositional tasks, HiMEM achieves 61.0\% success, exceeding the backbone and ECHO by 24.0 and 14.0 percentage points, respectively. Figure~\ref{fig:geometric_retrieval_example} further illustrates how hierarchical retrieval reuses experiences from different tasks to support unseen multi-step manipulation.







\subsection{Efficiency and Memory Scale}
\label{subsec:system_analysis_main}

We retain 25\%, 50\%, 75\%, and 100\% of the source execution pool and report segment counts, success, and cone-tree/flat-$k$NN retrieval latency in Table~\ref{tab:memory_scaling}. Source splits, query sets, hardware, and timing conventions are matched. Online latency is decomposed into backbone inference, tree retrieval, rule lookup, candidate aggregation, and residual application under a common measurement convention. Scene-condition refreshes are timed separately at task or subgoal transitions. Offline costs separately record video decoding and VLM analysis, evidence verification, tree updates, and rule consolidation, together with the number of processed episodes. At each memory scale, the measurements also track successful and failed leaves, rule counts, storage, semantic lookup, and end-to-end inference under matched device, precision, warm-up, and synchronization conventions, distinguishing retrieval latency from complete system cost.

Table~\ref{tab:memory_scaling} compares ECHO and HiMEM at matched source-pool fractions and reports each method's resulting episodic segment count, separating source-pool scale from the effect of memory pruning.

\begin{table}[t]
  \centering
  \caption{Three additional ablations assess virtual synthesis, failure evidence, and cross-episode consolidation.}
  \label{tab:cross_suite_generalization}
  \footnotesize
  \renewcommand{\arraystretch}{1.12}
  \setlength{\tabcolsep}{2.2pt}

  \begin{tabular}{@{}lccccc@{}}
    \toprule
    \multirow{2}{*}{Setting}
      & \multirow{2}{*}{Backbone}
      & \multicolumn{2}{c}{ECHO}
      & \multicolumn{2}{c}{HiMEM} \\
    \cmidrule(lr){3-4}
    \cmidrule(l){5-6}
      & & \shortstack{Held-out\\memory}
        & \shortstack{Same-target\\memory}
        & \shortstack{Held-out\\memory}
        & \shortstack{Same-target\\memory} \\
    \midrule

    \shortstack[l]{LIBERO\\cross-suite}
      & 80.70 & 89.31 & \textsc{93.48}
      & \textsc{92.78} & \textsc{95.30} \\

    \shortstack[l]{Real robot\\(G1--G5)}
      & 37.0 & 47.0 & --
      & 61.0 & \textsc{-} \\

    \bottomrule
  \end{tabular}
\end{table}

\begin{table}[t]
  \centering
  \caption{Memory-scale efficiency on LIBERO-Long. ECHO and HiMEM use matched source-pool fractions; segment counts reflect each method's resulting episodic memory. Tree and kNN report episodic retrieval latency (ms); other costs are measured separately.}
  \label{tab:memory_scaling}
  \label{tab:memory_efficiency}
  \footnotesize
  \renewcommand{\arraystretch}{1.08}
  \setlength{\tabcolsep}{2.2pt}
  \begin{tabular}{@{}clcccc@{}}
  \toprule
  \shortstack{Source\\pool} & Method & Segments
  & \shortstack{Success\\(\%)}
  & \shortstack{Tree\\(ms)}
  & \shortstack{kNN\\(ms)} \\
  \midrule

  \multirow{2}{*}{25\%}
    & ECHO & \textsc{692} & \textsc{91.67} & \textsc{4.9} & \textsc{7.8} \\
    & \cellcolor{gray!10}\textbf{HiMEM}
    & \cellcolor{gray!10}\textsc{420}
    & \cellcolor{gray!10}\textsc{92.25}
    & \cellcolor{gray!10}\textsc{4.4}
    & \cellcolor{gray!10}\textsc{4.8} \\

  \cmidrule[0.3pt](lr){1-6}

  \multirow{2}{*}{50\%}
    & ECHO & \textsc{1384} & \textsc{92.25} & \textsc{5.5} & \textsc{15.3} \\
    & \cellcolor{gray!10}\textbf{HiMEM}
    & \cellcolor{gray!10}\textsc{834}
    & \cellcolor{gray!10}\textsc{92.67}
    & \cellcolor{gray!10}\textsc{5.1}
    & \cellcolor{gray!10}\textsc{9.0} \\

  \cmidrule[0.3pt](lr){1-6}

  \multirow{2}{*}{75\%}
    & ECHO & \textsc{2077} & \textsc{93.25} & \textsc{5.8} & \textsc{22.1} \\
    & \cellcolor{gray!10}\textbf{HiMEM}
    & \cellcolor{gray!10}\textsc{1250}
    & \cellcolor{gray!10}\textsc{94.75}
    & \cellcolor{gray!10}\textsc{5.2}
    & \cellcolor{gray!10}\textsc{13.9} \\

  \cmidrule[0.3pt](lr){1-6}

  \multirow{2}{*}{100\%}
    & ECHO & \textsc{2769} & \textsc{93.50} & \textsc{6.1} & \textsc{29.5} \\
    & \cellcolor{gray!10}\textbf{HiMEM}
    & \cellcolor{gray!10}\textsc{1665}
    & \cellcolor{gray!10}\textsc{95.25}
    & \cellcolor{gray!10}\textsc{5.6}
    & \cellcolor{gray!10}\textsc{12.8} \\

  \bottomrule
\end{tabular}
\end{table}

To assess how episodic memory pruning affects task success, we fix the source experiences, consolidated semantic memories, policy checkpoint, and evaluation configurations, varying only the target episodic-vector pruning ratio (0\%, 20\%, 40\%, 60\%, and 80\%). Table~\ref{tab:pruning_ratio} reports the actual pruning ratio, retained episodic-vector count, LIBERO-Long success rate. Coverage constraints preserve non-redundant experiences, so actual pruning may fall below the target.

\begin{table}[t]
  \centering
  \caption{Effect of episodic memory pruning on LIBERO-Long with $\pi_0$ + HiMEM. Pruning ratios use the pre-pruning episodic-vector count as the denominator.}
  \label{tab:pruning_ratio}
  \footnotesize
  \renewcommand{\arraystretch}{1.10}
  \setlength{\tabcolsep}{2.4pt}
  \begin{tabular}{@{}cccc@{}}
    \toprule
    \shortstack{Target\\pruned (\%)} &
    \shortstack{Actual\\pruned (\%)} &
    \shortstack{Retained\\vectors} &
    \shortstack{Success\\(\%)} \\
    \midrule
    0  & 0    & 2769 & 95.3 \\
    20 & 13.1 & 2406 & 95.2 \\
    40 & 25.2 & 2071 & 94.9 \\
    60 & 42.8 & 1584 & 94.0 \\
    80 & 56.6 & 1202 & 92.9 \\
    \bottomrule
  \end{tabular}
\end{table}

%% file: sections/05_limitation_and_conclusion.tex
\section{Limitations}
\label{sec:limitations}

HiMEM relies on recorded execution videos and verified subgoal outcomes to extract reliable manipulation patterns across episodes. For newly encountered scenarios with limited historical evidence, meta-semantic rules may initially lack sufficient statistical support, while the system can still utilize its hierarchical episodic memory for retrieval and decision-making. Furthermore, video analysis and rule updates introduce additional offline computational overhead, but these operations are performed after task execution and do not interfere with online action generation.

\section{Conclusion}
\label{sec:conclusion}

In this paper, we proposed HiMEM, a hippocampus-inspired long-term memory framework for VLA models in complex 3D environments. To address memory noise and limited cross-task transfer, HiMEM introduces a semantic memory consolidation mechanism that extracts generalizable manipulation knowledge from episodic experiences. It further employs semantic-guided episodic memory pruning to reduce memory redundancy and a flexible memory composition mechanism that jointly utilizes episodic and semantic memories for complex manipulation tasks. Extensive experiments across multiple benchmarks and VLA backbones demonstrate the effectiveness of HiMEM in improving manipulation performance and cross-scenario generalization. These results highlight the importance of integrating episodic experiences with generalizable semantic knowledge for robotic manipulation.